\documentclass[letterpaper]{article} 
\usepackage[]{aaai24}  
\usepackage{times}  
\usepackage{helvet}  
\usepackage{courier}  
\usepackage[hyphens]{url}  
\usepackage{graphicx} 
\usepackage{natbib}  
\usepackage{caption} 
\usepackage{algorithm}

\usepackage{algorithmicx}
\usepackage{xspace}
\usepackage{subcaption}
\usepackage{enumitem}
\usepackage{amsmath}
\usepackage{amssymb}
\usepackage{algpseudocode}
\usepackage{array}
\usepackage{color}

\let\titleold\title
\renewcommand{\title}[1]{\titleold{#1}\newcommand{\thetitle}{#1}}
\def\maketitlesupplementary
   {
   \newpage
   \begingroup 
       \centering
       \Large
       \textbf{\thetitle}\\\vspace{0.5em}
       Supplementary Material\\\vspace{1.0em}
   \endgroup 
   }

\usepackage{newfloat}
\usepackage{listings}
\DeclareCaptionStyle{ruled}{labelfont=normalfont,labelsep=colon,strut=off} 
\floatstyle{ruled}
\newfloat{listing}{tb}{lst}{}
\floatname{listing}{Listing}
\newcommand{\sssec}[1]{\vspace*{0.05in}\noindent\textbf{#1}}

\def\sysname{{\textsc{SecDT}}}

\newcolumntype{M}[1]{>{\centering\arraybackslash}p{#1}}

\title{
Training on Fake Labels: Mitigating Label Leakage in Split Learning via Secure Dimension Transformation
}
\author{
    Yukun Jiang\equalcontrib \textsuperscript{\rm 1},
    Peiran Wang\equalcontrib \textsuperscript{\rm 2},
    Chengguo Lin \textsuperscript{\rm 3},
    Ziyue Huang \textsuperscript{\rm 1},
    Yong Cheng \textsuperscript{\rm 1}
}
\affiliations{
    \textsuperscript{\rm 1}Tencent Inc.\\
    \textsuperscript{\rm 2}Tsinghua University\\
    \textsuperscript{\rm 3}Peking University\\

}

\usepackage{bibentry}

\begin{document}

\nocopyright
\maketitle

\begin{abstract}
Two-party split learning has emerged as a popular paradigm for vertical federated learning. 
To preserve the privacy of the label owner, split learning utilizes a split model, which only requires the exchange of intermediate representations (IRs) based on the inputs and gradients for each IR between two parties during the learning process. 
However, split learning has recently been proven to survive label inference attacks. Though several defense methods could be adopted, they either have limited defensive performance or significantly negatively impact the original mission. 
In this paper, we propose a novel two-party split learning method to defend against existing label inference attacks while maintaining the high utility of the learned models. 
Specifically, we first craft a dimension transformation module, SecDT, which could achieve bidirectional mapping between original labels and increased $K$-class labels to mitigate label leakage from the directional perspective. 
Then, a gradient normalization algorithm is designed to remove the magnitude divergence of gradients from different classes. 
We propose a softmax-normalized Gaussian noise to mitigate privacy leakage and make our $K$ unknowable to adversaries. 
We conducted experiments on real-world datasets, including two binary-classification datasets (Avazu and Criteo) and three multi-classification datasets (MNIST, FashionMNIST, CIFAR-10); we also considered current attack schemes, including direction, norm, spectral, and model completion attacks. The detailed experiments demonstrate our proposed method's effectiveness and superiority over existing approaches. For instance, on the Avazu dataset, the attack AUC of evaluated four prominent attacks could be reduced by 0.4532±0.0127.
\end{abstract}
\section{Introduction}

\begin{figure*}
  \centering
  \includegraphics[width=0.9\textwidth]{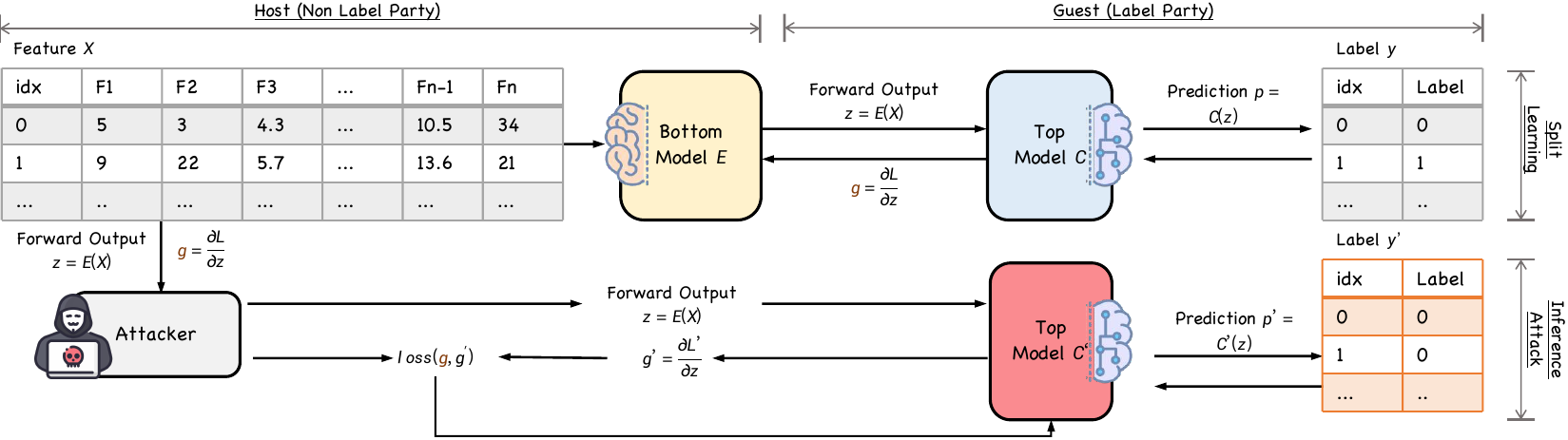}
  \caption{
  In the threat model, we assume the host to be the attacker who wants to get the private data label owned by the guest.
  This inference attack can be done through the inference attack from the backward gradient.
  }
  \label{fig:threat}
\end{figure*}

Deep learning has been applied in many areas of people's daily lives. 
However, the paradigm of data-centralized deep learning has been continuously questioned since people's concerns about their privacy rose.
For a typical scene, consider that an online shopping company A owns the clients' purchase records, while an online video website owner B has the clients' advertisement data. 
To learn how the advertisement on B will impact clients' purchase tendency on A.
A and B need to learn a global model using A's data labels and B's features. 
However, due to privacy concerns and regulations such as GDPR for clients' data, A and B cannot directly share their data to train a global model \cite{turina2021federated, thapa2021advancements, zhang2023privacy, turina2020combining}.

\par Split learning \cite{yang2019federated, gupta2018distributed, poirot2019split, vepakomma2018split, liu2024vertical} is a solution to such a scenario, allowing the feature and label owners to train a machine learning model jointly.
In split learning, the neural network's training process is split into the non-label and label parties \cite{langer2020distributed}. 
At the beginning of split learning, the two parties apply Private Set Intersection (PSI) protocols \cite{abuadbba2020can} to find the intersection of their data examples and establish the alignment of data example IDs. 
During training, the non-label party will use a bottom model to obtain the intermediate layer of its data examples and send it to the label party. 
Then, the label party will complete the rest of the neural network training process. 
The label party will apply a top model to predict the non-label party's data examples. 
Next, in the backpropagation process, the label party computes the gradient from its loss function and sends the parameter gradient back to the non-label party. 
Thus, the non-label party can take the parameter gradient from the label party to train its bottom model.
Split learning has been applied in many areas \cite{pham2023binarizing, abuadbba2020can, lin2024split, wu2023split, matsubara2022split, samikwa2022ares, wu2023split_2, park2021federated}.

\par Though the goal of split learning is to preserve privacy for both parties, the challenge of privacy leakage still exists within split learning. The non-label party can establish a gradient inversion attack \cite{kariyappa2021gradient}, a label leakage attack that allows an adversarial input owner to learn the label owner’s private labels by exploiting the gradient information obtained during split learning. In the gradient inversion attack, the label leakage attack is treated as a supervised learning problem by developing a novel loss function using specific key properties of the dataset and model parameters instead of labels. The non-label party can derive the labels of data examples via a gradient inversion attack, thus harming the privacy of the label party. Split learning urgently requires a protection mechanism to guard the privacy of label parties. In this paper, we propose these contributions to achieve this goal:

\begin{itemize}[noitemsep,topsep=1pt, leftmargin=*]
    \item We craft a dimension transformation module that could achieve bidirectional mapping between original and increased $K$-class labels to mitigate label leakage from the directional perspective. 
    \item A gradient normalization algorithm is designed to remove the magnitude's divergence of gradients w.r.t. samples from different classes.
    \item We propose one softmax-normalized Gaussian noise to mitigate privacy leakage and make our $K$ unknowable to adversaries.
\end{itemize}
\section{Background and Related Work}

\subsection{Privacy Leakage in Split Learning}
Though in split learning, the label party and non-label party will only share intermediate representation, privacy leakage threats still exist in this system \cite{geiping2020inverting, jin2021cafe, wu2022coupled, jiang2022vf, qi2022fairvfl}.

\sssec{Feature leakage}.
First, the forward intermediate representations can leak the private feature information in the non-label party.
\cite{vepakomma2019reducing} is the first to such feature leakage in split learning and provided a defense solution using distance correlation.
By creating a novel loss function employing specific key properties of the dataset and model parameters, \cite{kariyappa2021gradient} created a Gradient Inversion Attack (GIA), which converts the attack into a supervised learning problem.

\sssec{Label leakage}.
Second, the backward intermediate representation can leak the private label information in the label party.
\cite{li2022label} propose a norm-based method for leaking private labels in the conversion prediction problem. 
Their method is inspired by the high-class imbalance of the training dataset for the conversion prediction task.
Due to this imbalance, the gradients' magnitude is more significant when the unusual class is encountered. So, an adversarial input owner can infer the private class labels by considering the gradients' norm. Instead, they investigate whether label information may be disclosed through backward interaction from the label party to the non-label party.
\cite{fu2022label} discover that a malicious participant can exploit the bottom model structure and the gradient update mechanism to gain the power to infer the privately owned labels. Worse still, by abusing the bottom model, he/she can infer labels beyond the training dataset. Based on their findings, they propose a set of novel label inference attacks.
\cite{sun2022label} uses SVD to find correlations between embeddings and labels. Analyzing the mean and singular vectors assigns scores, clusters samples, and infers labels, bypassing privacy protections.
\cite{liu2021batch} first investigate the potential for recovering labels in the vertical federated learning context with HE-protected communication and then demonstrate how training a gradient inversion model can restore private labels. Additionally, by directly substituting encrypted communication messages, they demonstrate that label-replacement backdoor attacks may be carried out in black-boxed VFL (termed "gradient-replacement attack").

\par Our work \sysname\ focuses on defending label leakage threats. 

\subsection{Privacy Protection in Split Learning}

Techniques to protect communication privacy in FL generally fall into three categories: 1) cryptographic methods such as secure multi-party computation \cite{bonawitz2017practical, zhang2020batchcrypt, wan2024misa, pereteanu2022split}; 2) system-based methods including trusted execution environments \cite{subramanyan2017formal}; and 3) perturbation methods that shuffle or modify the communicated messages \cite{abadi2016deep, mcmahan2017communication, zhu2019deep, erlingsson2019amplification, cheu2019distributed, cheng2021secureboost}. Defense against label leakage threats can be categorized into the 3rd category.

Previous researchers have proposed several defense schemes to defend against label leakage threats.
\cite{li2022label} propose Marvell, which strategically determines the form of the noise perturbation by minimizing the label leakage of a worst-case adversary. The noise is purposefully designed to reduce the gap between the gradient norms of the two classes, which deters the attack. However, it requires a large amount of extra computation that slows down the speed of split learning.
\cite{abadi2016deep} proposed using differential privacy in deep learning to protect privacy. We utilize differential privacy in vertical federated learning on the guest side to safeguard the confidentiality of data labels.
\cite{wei2023FedAds} propose MixPro, an innovative defense mechanism against label leakage attacks in Vertical Federated Learning (VFL) that employs a two-step approach: Gradient Mixup and Gradient Projection.
To prevent the non-label party from assuming the genuine label, \cite{liu2021batch} also introduced "soft fake labels." To increase confusion, a confusional autoencoder (CoAE) is used to create a mapping that transforms one label into a soft label with a more significant likelihood for each alternative class.

\subsection{System Cost of Split Learning}

Other than privacy issues, previous researchers have also investigated the system cost of split learning. To overcome the training process's sequential nature, split learning extensions have been proposed \cite{jeon2020privacy, thapa2022splitfed, turina2020combining, abedi2024fedsl, guo2021lightfed, shen2023ringsfl, maini2022characterizing}. More prominently, in \cite{turina2020combining}, split learning is combined with federated learning (i.e., SplitFed learning) to yield a more scalable training protocol. Here, the server handles the forward signal of the clients’ network in parallel (without aggregating them) and updates the weights. The clients receive the gradient signals and update their local models in parallel. Then, they perform federated learning to converge to a global function before the next iteration of split learning. This process requires an additional server that is different from the one hosting. Split learning gained particular interest due to its efficiency and simplicity. Namely, it reduces the required bandwidth significantly compared to other approaches, such as federated learning \cite{singh2019detailed, vepakomma2018no, oh2022locfedmix, koda2020communication}. Indeed, for large neural networks, intermediate activation for a layer is consistently more compact than the network’s gradients or weights for the entire network.
Furthermore, the clients' computational burden is smaller than that caused by federated learning. Indeed, clients perform forward/backward propagation on a small portion of the network rather than the whole. This allows split learning to be successfully applied to the Internet of Things (IoT) and edge-device machine learning settings \cite{gao2020end, koda2019one, wu2023split, zhang2023privacy, han2021accelerating}.
\section{Label Inference Attack}\label{sec:label_infer}

In this section, we first show the system model of split learning and threat model of label inference attacks, followed by five kinds of label inference attacks.

\subsection{System Model}\label{sec:label_infer:system}

In two-party split learning, considering a training dataset $\{X_{i}, y_{i}\}_{i=1}^{N}$, the host (non-label party) has the features $\{X_{i}\}_{i=1}^{N}$ and the guest (label party) has the corresponding labels $\{y_{i}\}_{i=1}^{N}$ as shown in Figure \ref{fig:threat}. 
They jointly train a model based on the training dataset $\{X_{i}, y_{i}\}_{i=1}^{N}$ without leaking $\{X_{i}\}_{i=1}^{N}$ and $\{y_{i}\}_{i=1}^{N}$ to each other. 
Specifically, the host trains a bottom model $E$ while the guest trains a top model $C$. 
In each iteration, the host sends the forward embedding output $z=E(X)$ of the bottom model $E$ to the guest, then the guest will use $z$ as the input of the top model $C$ and send the backward gradient $g$ to the host. 
After receiving the gradient $g$, the host uses $g$ to update the bottom model code $E$. 
The host will not access the label $\{y_{i}\}_{i=1}^{N}$, while the guest will not access the data feature $\{X_{i}\}_{i=1}^{N}$, thus protecting the privacy of both the host and the guest.


\subsection{Potential Attacks}\label{sec:label_infer:attack}

\par Then we consider the malicious host who wants to get the label held by the guest. We consider five types of attacks in our threat model:

\sssec{Direction attack \cite{li2022label}} For a given example, all examples of the same class provide positive cosine similarity, but all examples of the opposite class produce negative cosine similarity. The non-label party can identify the label of each case if the problem is class-imbalanced and knows there are more negative examples than positive ones. The class is negative if more than half of the examples produce positive cosine similarity; otherwise, it is positive. In many real-world scenarios, the non-label party can reasonably guess which class contains most examples in the dataset without ever having any access to the data. For instance, in disease prediction, the prevalence of a given disease in the general population is almost always much lower than 50\%.

\sssec{Norm attack \cite{li2022label}} The model tends to be less confident during training that a positive example will be positive than a negative example will be negative. Moreover, for both the positive and negative cases, the norm of the gradient vector $||g||_2$ is on the same order of magnitude (has an equivalent distribution). As a result, the gradient norm $||g||_2$ for positive examples is typically higher than for negative instances. The scoring function $r_n(g)=||g||_2$ is a good predictor of the hidden label $y$. The privacy loss (leak AUC) compared to the attack $r_n$ is known as the norm leak AUC.

\sssec{Spectral attack \cite{sun2022label}.} The spectral attack utilizes singular value decomposition (SVD) to exploit the correlation between intermediate embeddings and private labels in machine learning models. By calculating the mean and top singular vector of the embeddings, the attack assigns scores to each sample, clusters them, and then infers the labels based on the distribution of these scores, effectively stealing private labels despite the presence of certain privacy-preserving techniques.

\sssec{Model competition attack \cite{fu2022label}.} The model competition attack involves an adversary obtaining a trained bottom model, adding random inference layers to form a complete model, and then fine-tuning it with a small set of auxiliary labeled data. The adversary refines the model using semi-supervised learning algorithms tailored to the dataset's domain. The result is a fully-trained model that can predict labels for any data, allowing the adversary to infer labels for any sample of interest while adhering to the rules of VFL, thus conducting a passive attack without active interference.

\section{The Proposed \sysname}\label{sec:scheme}

\begin{figure*}[t]
\centering
\includegraphics[width=0.9\textwidth]{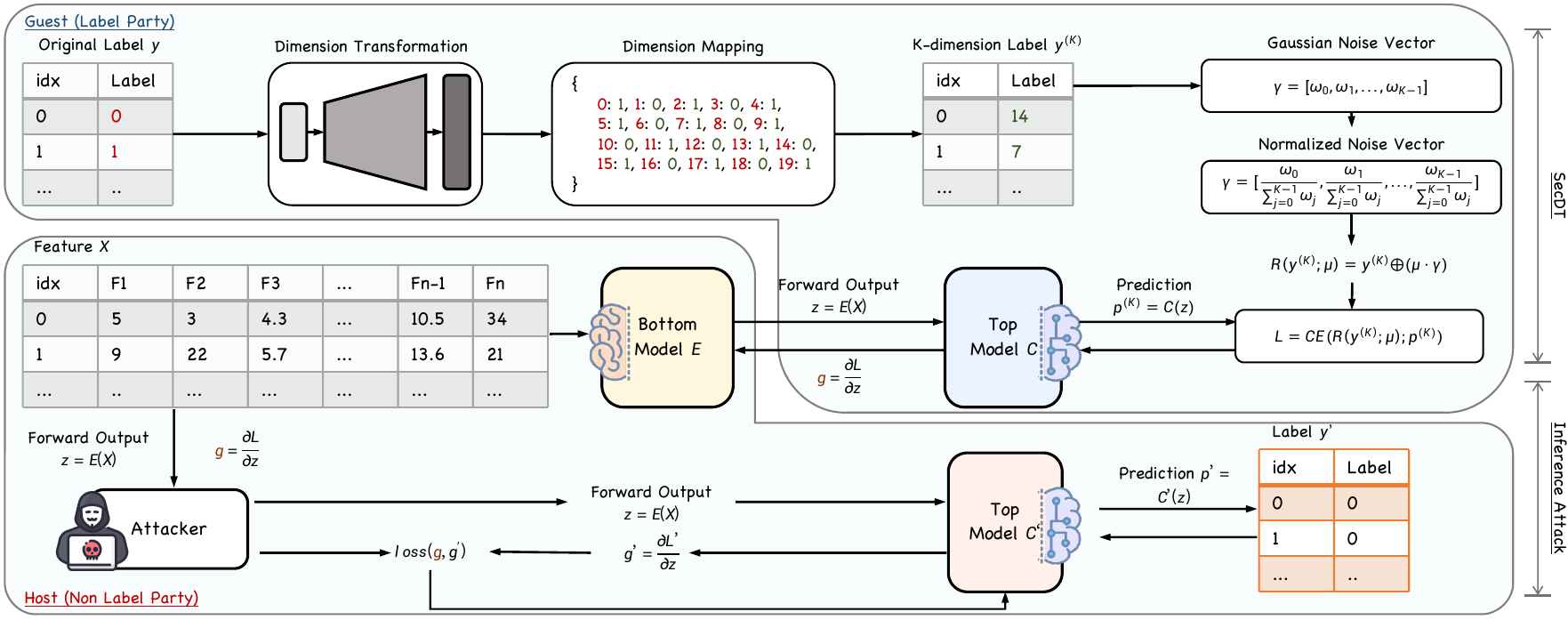}
\caption{
This figure outlines the SecDT algorithm's training process, which includes dimension transformation to expand the label space, gradient normalization to standardize gradient magnitudes, and noise-based randomization to introduce uncertainty into the label dimensions.
}
\label{fig:overview}
\end{figure*}

\subsection{Overview}
Figure \ref{fig:overview} shows the training workflow of our proposed \sysname, which comprises three parts, i.e., dimension transformation, gradient normalization, and randomization. 
In two-party split learning, there is a host (feature party) and a guest (label party), but only the guest has the training labels. 
However, the private label information could be inferred by the offensive host from the backward gradients $\{\mathbf{g}_{i}\}_{i=1}^{N}$. 
Generally speaking, existing label inference attacks against split learning are designed from two perspectives: direction and magnitude~\cite{li2022label, sun2022label, fu2022label}. 
Therefore, the dimension transformation is crafted to achieve bidirectional mapping between original and increased $K$-class labels, mitigating label leakage from the directional perspective. 
Then, to remove the magnitude's divergence of gradients, we design a magnitude normalization algorithm to achieve this goal. 
Furthermore, we introduce a randomization module in which two random noises are proposed to mitigate privacy leakage and make our $K$ unknowable to adversaries. 
All the defenses are conducted by the guest, which is invisible to the host. 

\subsection{Dimension Transformation}\label{sec:scheme:dim}
It is intuitive that for the adversaries, the original classification tasks are more vulnerable to label inference attacks than the multi-class classification tasks, especially from the directional perspective. 


\sssec{Dimension-increased transformation} Could we transform original labels to multi-class labels during model training? 
To achieve this goal, we craft the dimension transformation module. 
Transfer from binary to multi-class:
Specifically, given dataset $D = \{X_i,y_i\}_{i=1}^N$ with labels $\{y_{i}\}_{i=1}^{N}$, we have
\begin{equation}
    \text{One-hot}(y_i) = \hat{y}_{i} \in \text{One-hot}(\{0, \cdots, k-1\}),
\end{equation}
where One-hot$(\cdot)$ converts numbers $0$ to $k-1$ to one-hot encodings.
For instance, if $k=3$ (i.e., the task is a three-category task), we have $\text{One-hot}(0, 1, 2) = \{[1,0,0], [0,1,0], [0,0,1]\}$.
Then, given labels $\{y_{i}\}_{i=1}^{N}$ and targeted (increased) dimension $K$, we define a mapping $M_{k,K}$ from $k$-dimension to $K$-dimension as
\begin{equation}
    \mathcal{M}_{k,K}\left(y_{i}\right) = m_{K,y_i},
\end{equation}
where $m_{K,y_i}$ are elements that randomly selected from corresponding mapping pool $\rho_{K,y_i}$ for label $y_i$.
To generate the required mapping pools $\rho_{K,y_i}$, with specific $K$, an ordered set of one-hot encodings is defined as
\begin{equation}
    \Psi = \{\text{One-hot}(0), \cdots, \text{One-hot}(K-1)\}.
\end{equation}
Then, we shuffle $\Psi$ randomly or according to specific rules to get 
\begin{equation}
     \Psi_{s} = \text{Shuffle}(\Psi)
\end{equation}
and separate $\Psi_{s}$ into $k$ disjoint mapping pools given by
\begin{equation}
    \rho_{K,0}, \cdots, \rho_{K, k-1},
\end{equation}
where $\bigcap_{y=0}^{k-1} \rho_{K,y} = \emptyset$ and $\bigcup_{y=0}^{k-1} \rho_{K,y} = \Psi_{s}$.
There could be many rules for dividing $\Psi_{s}$ to $\rho_{K,y}$. For simplicity, in this work, unless otherwise mentioned, we divide $\Psi_{s}$ equally to $\rho_{K,0}, \cdots, \rho_{K, k-1}$.
Indicating the length (number of elements) of $\rho_{K, y}$ by $\sigma_y = \frac{K}{k}$, for each $\rho_{K, y}$, we have
\begin{equation}
    \rho_{K, y} = \{ \tau_{y, 0}, \cdots, \tau_{y, \sigma_y - 1} \}.
\end{equation}

Then, we have the dataset $D_K = \{X_i,\mathcal{M}_{k,K}\left(y_{i}\right)\}_{i=1}^N$ with increased the $K$-dimension labels to optimize the objective
\begin{equation}
\label{equation:loss_K}
    \min_{E, C} \mathcal{L}(\theta, D_K) = \min_{E, C} \frac{1}{N} \sum_{i=1}^N L\left(C(E(X_i)), \mathcal{M}_{k,K}(y_i)\right),
\end{equation}
where $L(\cdot, \cdot)$ is for computing the cross entorpy.
Because dimension transformation works before training of the model and the dimension-increased labels are fixed, optimizing the Equation \ref{equation:loss_K} could also optimize the objective
\begin{equation}
    \min_{E, C} \mathcal{L}(\theta, D) = \min_{E, C} \frac{1}{N} \sum_{i=1}^N L\left(E(C((X_i)), \hat{y}_i\right).
\end{equation}

\sssec{Dimension-decreased transformation}
The $K$-dimension prediction could not be directly used as the final result during inference time. 
Intuitively, with the $K$-dimension prediction $p^{(K)} = C(z) = C(E(X))$, we could derive original $k$-dimension inference result $p$ based on a \underline{m}aximum \underline{m}apping function $\mathcal{MM}(\cdot)$ given by  
\begin{equation}
    p = \mathcal{MM}(p^{(K)}) = y, \ if \ p^{(K)} \in \rho_{K, y}.
\end{equation}

However, as turning a $k$-classification task into a $K$-classification task will increase the difficulty of model training, the aforementioned function $\mathcal{MM}(\cdot)$ results in compromised model performance compared with the original $k$-classification task.
We believe this is because the data whose labels belong to the same mapping pool $\rho_{K, y}$ essentially have similar characteristics.
\emph{We believe this is because the data samples used for inference are similar but not identical to the training samples, resulting in the $p^{(K)}$ to be linear combinations of one-hot encodings in $\rho_{K, y}$ for feature $X$.}
To overcome this challenge, in our \sysname, we propose to realize performance-reserved dimension-decreased transformation based on a novel \underline{w}eighted \underline{m}apping function $\mathcal{WM}(\cdot)$ given by
\begin{equation}
    p = \mathcal{WM}(p^{(K)}) = {\arg \max}_{0 \leq y \leq k-1} \mathbf{w}_y \cdot p^{(K)},
\end{equation}
where $P^{(K)}$ is the weight.
Here, $\mathbf{w_{y}} \cdot p^{(K)}$ is the inner product of $\mathbf{w}_y$ and $\cdot p^{(K)}$, where $\mathbf{w}_y$ represents the result of element-wise addition ($\oplus$) of one-hot encodings in the mapping pool $\rho_{K, y}$ that
\begin{equation}
    \mathbf{w_{y}} = \tau_{y, 0} \oplus \tau_{y, 1} \oplus \cdots \oplus \tau_{y, \sigma_y - 1}.
\end{equation}
With function $\mathcal{WM}(\cdot)$, we covert the linear combination of one-hot encodings in $\rho_{K, y}$ to the inner product as the confidence that $P^{(K)}$ belongs to label $y$.
Our results show that the proposed weighted mapping significantly guarantees the effectiveness of the task model.

\subsection{Gradient Normalization}\label{sec:scheme:norm}
Inspired by \cite{cao2021fltrust} that normalizes gradients from suspicious clients in horizontal federated learning, in \sysname, we make the first attempt to normalize gradients in the cut layer of split learning, which could fully avoid label inference attacks conducted based on the gradients' magnitude. 
\cite{cao2021fltrust} normalize gradients with a trust-worthy gradient locally computed by the server, but there is no single trust-worthy gradient in our \sysname.
Hence, since all gradients in our \sysname are clean (i.e., have not been manipulated by adversaries), literally, each gradient could be used to normalize others. 
In our \sysname, the minimum, mean, and maximum $\ell_{2}$-norm among all gradients in the current mini-batch could be used to realize normalization.
In this paper, unless otherwise mentioned, the mean $\ell_{2}$-norm is considered as the \emph{standard norm} for normalization.
Specifically, when the batch size is set to be $B$, during each iteration, $B$ gradients $\{g_{b}\}_{b=1}^{B}$ are computed by the guest. 
Then, we normalize these gradients as
\begin{equation}
    \bar{g}_{b} = g_{b} \cdot \frac{\varphi}{\Vert g_{b} \Vert}, 
\end{equation}
where $\Vert \cdot \Vert$ represents $\ell_{2}$-norm and $\varphi$ is the selected standard $\ell_{2}$-norm. For instance, considering the mean $\ell_{2}$-norm as the standard, we have
\begin{equation}
    \varphi = \frac{1}{B}\sum_{b=1}^{B}\Vert g_{b} \Vert.
\end{equation}

\subsection{Noise-based Randomization}\label{sec:scheme:noise}
Considering potential adaptive attacks against our proposed \sysname, which may succeed after an attacker could infer our mapping pools $\rho_{K, y}$.
Hence, we aim to keep our increased dimension $K$ confidential to adversaries, making mapping pools unknowable.
In this work, we propose adding Softmax-normalized Gaussian noise (SGN) to make our increased dimension $K$ agnostic to adversaries.
Due to the task independence of introduced noise, the noise could also mitigate privacy leakage of split learning to existing attacks.
We assume that during \sysname's model training phase, each sample's label (target) is a $K$-dimension one-hot encode $\tau = [\zeta_{1}, \zeta_{2}, \cdots, \zeta_{K}]$. 
Then, we propose two noises as follows.

\sssec{Softmax-normalized Gaussian noise} 
For each sample's one-hot encoded label $\hat{y}$, the guest generates a noise vector
\begin{equation}
    \gamma = [\omega_{0}, \omega_{1}, \cdots, \omega_{K-1}],
\end{equation}
where $\omega$ follows standard Gaussian distribution, i.e., $\omega \sim \mathcal{N}(0, 1)$. Moreover, to make this noise more controllable, the guest normalizes $\gamma$ based on the softmax function as
\begin{equation}
\begin{split}
    \bar{\gamma} = & \text{Softmax}(\gamma) \\
    = & [ \frac{e^{\omega_{0}}}{\sum_{j=0}^{K-1}e^{\omega_{j}}},  \frac{e^{\omega_{1}}}{\sum_{j=0}^{K-1}e^{\omega_{j}}}, \cdots, \frac{e^{\omega_{K-1}}}{\sum_{j=0}^{K-1}e^{\omega_{j}}}]
\end{split}
\end{equation}


\sssec{Adding noise} With generated noise for each sample's label $\tau$, we add them into the initial label to obtain 
\begin{equation}
    \tau = \tau \oplus (\mu \cdot \gamma), 
\end{equation}
where $\mu$ is used to determine the noise level.


\section{Evaluation Setup}

\begin{figure*}[htbp]
  \centering
    \subfloat{\includegraphics[width=0.6\textwidth]{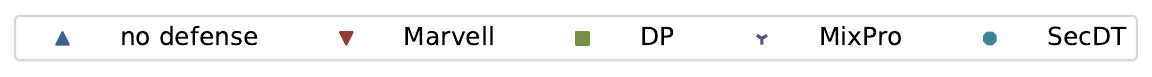}}
    \\
    \addtocounter{subfigure}{-1}
    \subfloat[{Avazu \& Norm attack}]{\includegraphics[width=0.33\textwidth]{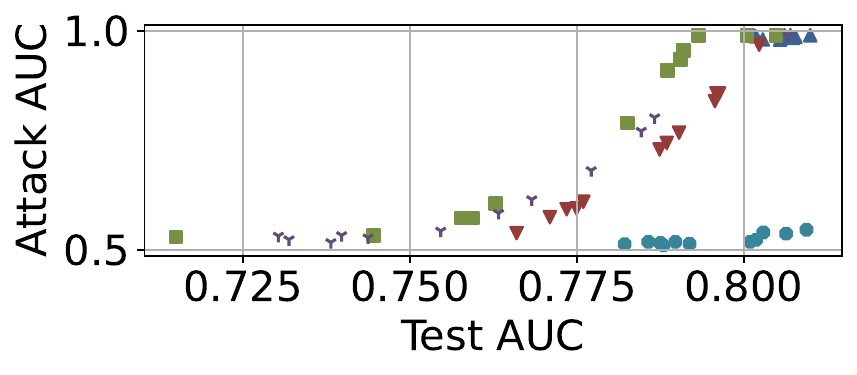}}
    \subfloat[{Avazu \& Direction attack}]{\includegraphics[width=0.33\textwidth]{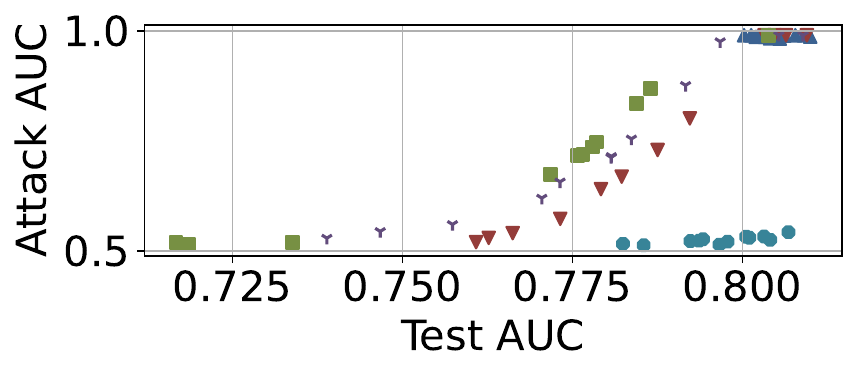}}
    \subfloat[{Avazu \& Spectral attack}]{\includegraphics[width=0.33\textwidth]{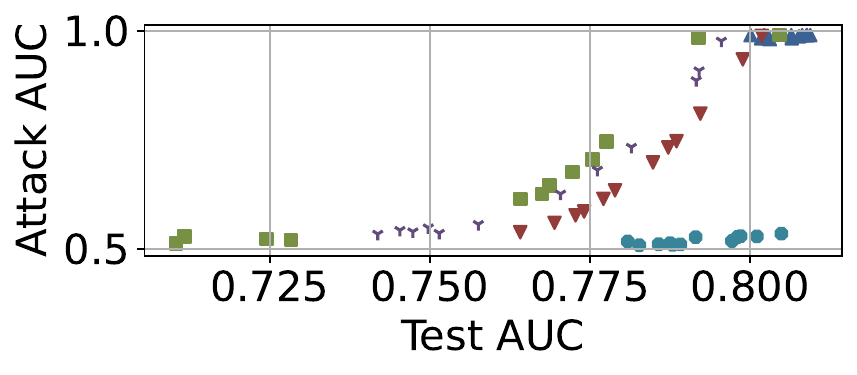}}
    \\
    \subfloat[{Criteo \& Norm attack}]{\includegraphics[width=0.33\textwidth]{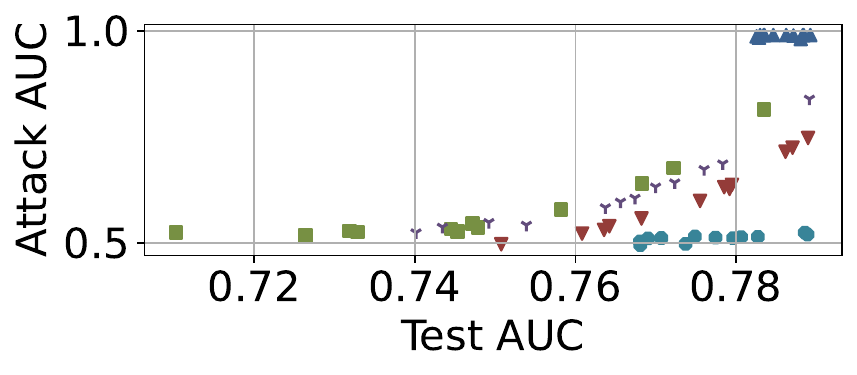}}
    \subfloat[{Criteo \& Direction attack}]{\includegraphics[width=0.33\textwidth]{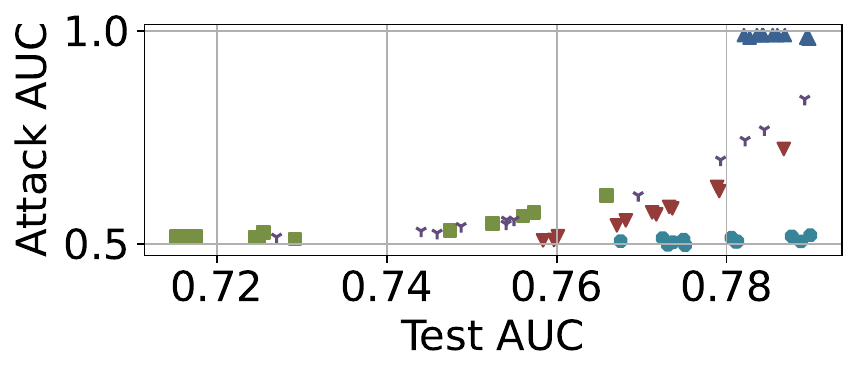}}
    \subfloat[{Criteo \& Spectral attack}]{\includegraphics[width=0.33\textwidth]{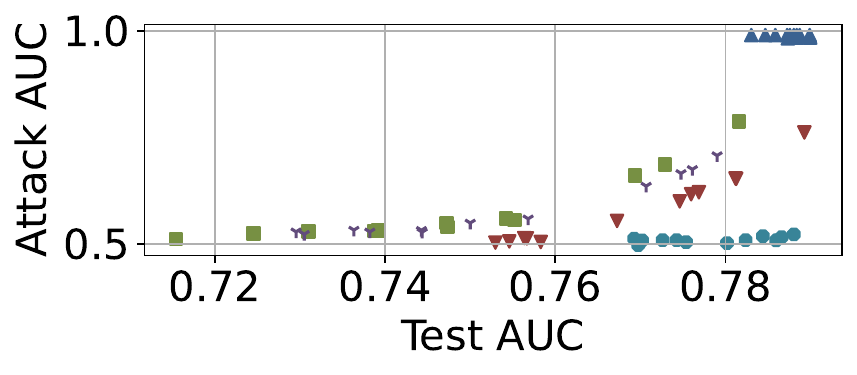}}
    \\
  \caption{
    This figure presents the results of SecDT's effectiveness against label inference attacks compared to other defense schemes such as Marvell, DP, and MixPro. It shows SecDT's superior performance in reducing Attack AUC without compromising Test AUC.
  }
  \label{fig:eval:sota:2}
\end{figure*}

In this section, we first discuss the datasets (\S\ref{sec:setup:dataset}), model architecture, and environment (\ref{sec:setup:envioronment}) for evaluation.
Then, we introduce the evaluated attacks (\S\ref{sec:setup:attacks}) and the compared schemes (\S\ref{sec:setup:schemes}) in our evaluation.

\subsection{Datasets}\label{sec:setup:dataset}

We selected three image datasets for the multiple classifications and two click prediction datasets for the binary classification:
\begin{itemize}[noitemsep,topsep=1pt, leftmargin=*]
    \item \textbf{Criteo} \cite{dataset:criteo}: Criteo is a CTR dataset provided by Criteo. The training set consists of some of Criteo's traffic over seven days. Each row corresponds to a display ad served by Criteo. 
    \item \textbf{Avazu} \cite{dataset:avazu}: Avazu is one of the leading mobile advertising platforms globally. It consists of 10 days of labeled click-through data for training and one day of ads data for testing (yet without labels). Only the first ten days of labeled data are used for benchmarking.
    \item \textbf{MNIST} \cite{dataset:mnist}: MNIST is a dataset of 70,000 28x28 pixel grayscale images of handwritten digits, split into 60,000 for training and 10,000 for testing.
    \item \textbf{FashionMNIST} \cite{dataset:fashion}: FashionMNIST is a 10-category dataset of 70,000 fashion item images, serving as a variation of MNIST for image classification tasks in machine learning.
    \item \textbf{CIFAR-10} \cite{dataset:cifar}: CIFAR-10 is a dataset with 60,000 32x32 color images across ten object classes for benchmarking image classification models.
\end{itemize}

Furthermore, we used AUC as the metric for the binary classification tasks, while we used accuracy as the metric for the multiple classification tasks.

\subsection{Experiment Envioronment}\label{sec:setup:envioronment}

For the three multiple-image classification datasets, we applied a CNN for classification. We used a WideDeep neural network to evaluate the two binary click classification datasets.
All experiments are performed on a workstation equipped with Intel(R) Xeon(R) CPU E5-2650 v4 @ 2.20GHz, 32GB RAM, and four NVIDIA RTX 3060 GPU cards. We use PyTorch to implement DNNs.

\subsection{Evaluated Attacks}\label{sec:setup:attacks}
In our experiment, we evaluated four types of label inference attacks:
\begin{itemize}[noitemsep,topsep=1pt, leftmargin=*]
    \item\textbf{Direction attack \cite{li2022label}}. When a non-labeled party knows the class imbalance, it can identify labels based on cosine similarity. If more negative examples exist, a higher proportion of positive similarity suggests a negative class, and vice versa.
    \item\textbf{Norm attack \cite{li2022label}}. The model's confidence in positive predictions is lower than in negative ones. The gradient norm $||g||_2$ for positive examples is usually higher, making it a good predictor for the hidden label $y$. This attack's privacy loss is measured by the norm leak AUC.
    \item\textbf{Spectral attack \cite{sun2022label}}. This attack uses SVD to find correlations between embeddings and labels. Analyzing the mean and singular vectors assigns scores, clusters samples, and infers labels, bypassing privacy protections.
    \item\textbf{Model competition attack \cite{fu2022label}}. An adversary obtains a bottom model, adds layers, and fine-tunes it with labeled data. Using semi-supervised learning, they create a model that can predict any data label, enabling passive inference without active interference.
\end{itemize}

\begin{figure*}[htbp]
  \centering
    \subfloat{\includegraphics[width=0.6\textwidth]{Figures/evaluation_figures/1-compare/compare_legend.pdf}}
    \\
    \addtocounter{subfigure}{-1}
    \subfloat[{Avazu}]{\includegraphics[width=0.5\textwidth]{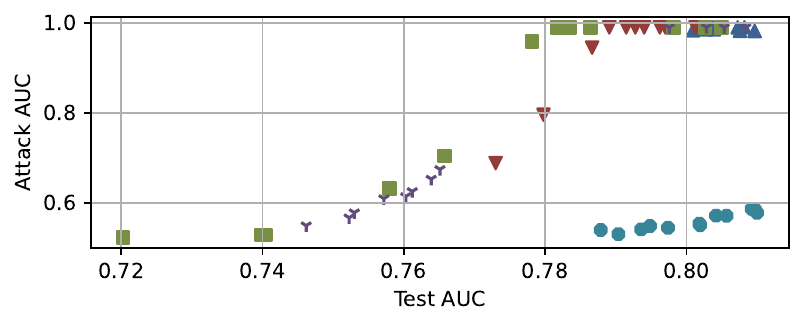}}
    \subfloat[{Criteo}]{\includegraphics[width=0.5\textwidth]{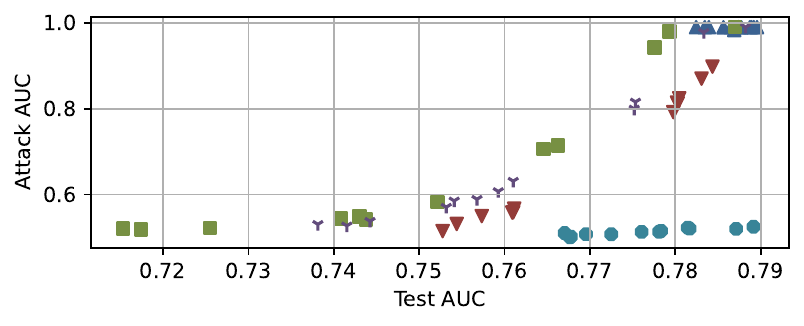}}
    \\
    \subfloat[MNIST]{\includegraphics[width=0.33\textwidth]{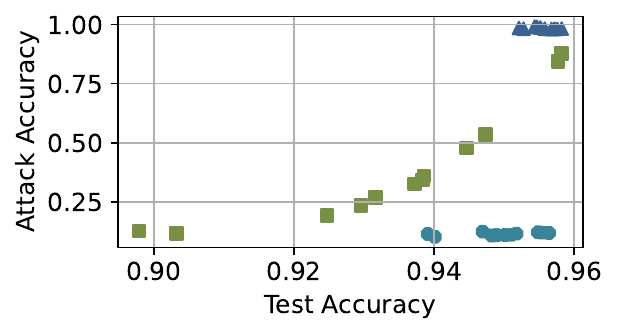}}
    \subfloat[{FashionMNIST}]{\includegraphics[width=0.33\textwidth]{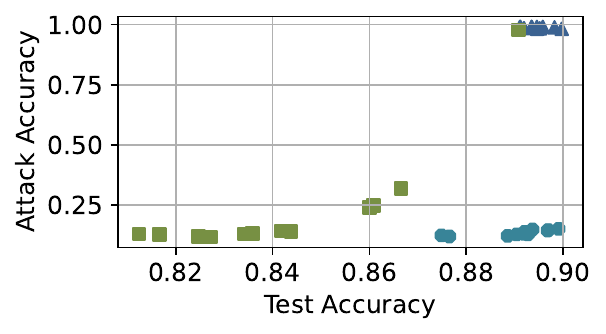}}
    \subfloat[{CIFAR}]{\includegraphics[width=0.33\textwidth]{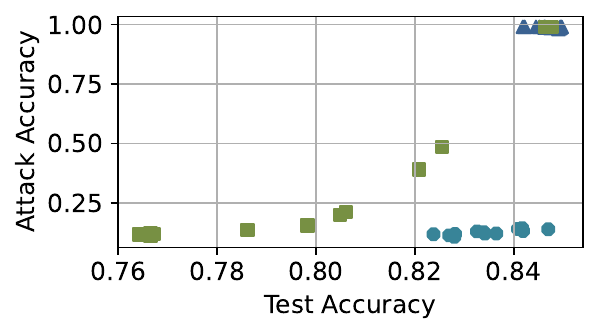}}
    \\
  \caption{
    This figure evaluates \sysname's performance against model completion attacks, where an adversary obtains a bottom model and fine-tunes it with auxiliary labeled data. \sysname\ demonstrates robustness against such attacks, with less impact on performance than other defense schemes.
  }
  \label{fig:eval:sota:10}
\end{figure*}

\subsection{Evaluated Schemes}\label{sec:setup:schemes}
We evaluated five label inference attack defense schemes in our experiments:

\begin{itemize}[noitemsep,topsep=1pt, leftmargin=*]
    \item \textbf{No defense}. The pure classification task in split learning without any defense mechanisms.
    \item \textbf{Marvell}. \cite{li2022label} proposed a random perturbation technique, which strategically finds the structure of the noise perturbation by minimizing the amount of label leakage (measured through our quantification metric) of a worst-case adversary (called Marvell).
    \item \textbf{DP}. \cite{abadi2016deep} proposed using differential privacy in deep learning to protect privacy. We utilize differential privacy in vertical federated learning on the guest side to protect the privacy of data labels.
    \item \textbf{MixPro}. MixPro \cite{wei2023FedAds} is an innovative defense mechanism against label leakage attacks in Vertical Federated Learning (VFL) that employs a two-step approach: Gradient Mixup and Gradient Projection. 
    \item \textbf{\sysname}. We proposed a dimension transformation method called \sysname, which transforms the classification task into a fake label classification task on the guest side.
\end{itemize}

\section{Evaluation Results}\label{sec:eval_results}


The evaluation results, depicted in Figure \ref{fig:eval:sota:2} and Figure \ref{fig:eval:sota:10}, provide a comparative analysis of \sysname\ against other state-of-the-art defense mechanisms. The Test AUC (Area Under the Curve) represents the model's performance on legitimate tasks, while the Attack AUC represents the model's utility for adversarial tasks. The balance between these two metrics is crucial in determining the effectiveness of a defense mechanism.

The results indicate that \sysname\ outperforms other defense schemes such as Marvell, DP (Differential Privacy), and MixPro in defending against label inference attacks without significantly compromising the model's accuracy. Specifically, \sysname\ demonstrated lower Attack AUC scores across norm, direction, and spectral attacks on the Avazu and Criteo datasets, signifying its superior ability to resist adversarial attacks.

Model completion attacks were also evaluated, involving an adversary obtaining a bottom model and fine-tuning it with auxiliary labeled data. The results show that \sysname's performance is less affected by this attack than other defense schemes. This indicates that \sysname's multi-faceted approach provides a more robust defense against various adversarial strategies.

\sssec{Impact of dimension transformation}.
A key aspect of \sysname's design is the dimension transformation module, which increases the complexity of the label space. The evaluation results suggest that this approach effectively increases adversaries' difficulty inferring labels. The transformation's impact is evident in the reduced Attack AUC scores, indicating that the increased dimensionality is a strong deterrent against label inference.

\sssec{Gradient normalization.}
The gradient normalization technique implemented in \sysname\ significantly mitigates attacks that rely on gradient magnitudes. The evaluation results show that normalization helps maintain a balanced Test AUC while significantly reducing the Attack AUC, particularly for norm attacks. This highlights the importance of gradient normalization in preserving the privacy of the label party.
\section{Conclusion}
In this study, we present a novel two-party split learning strategy, \sysname, to counter label inference attacks that have already been made. To reduce label leakage from a directional perspective, we first create a dimension transformation module that could achieve bidirectional mapping between binary and enhanced $K$-class labels. Then, a gradient normalization algorithm is created to eliminate the amplitude of gradients from various classes diverging. We also provide two random noises to prevent privacy leaks and prevent attackers from knowing our $K$. Studies on two real-world datasets show how successful and superior our suggested solution is to other ways.

\newpage
\bibliography{reference}
\clearpage
\newpage

\newpage
\appendix

\setcounter{section}{0} 
\maketitlesupplementary

\section{Algorithm Description of \sysname}

\begin{algorithm}
\caption{Training on Fake Labels: SecDT Algorithm}
\begin{algorithmic}[1]
\State \textbf{Input:} Dataset $D = \{X_i, y_i\}_{i=1}^N$, Desired dimension $K$, Noise level $\mu$

\Procedure{Dimension-Increased Transformation}{$D, K$}
    \State $\hat{y}_i \gets \text{One-hot}(y_i)$ \Comment{Convert labels to one-hot encoding}
    \State $\Psi \gets \{\text{One-hot}(0), ..., \text{One-hot}(K-1)\}$
    \State $\Psi_s \gets \text{Shuffle}(\Psi)$
    \State $\rho_{K,y} \gets \text{Partition}(\Psi_s, y)$ \Comment{Create mapping pools}
    \State $M_{k,K}(y_i) \gets \text{Map}(\hat{y}_i, \rho_{K,y_i})$ \Comment{Map to increased K-dimension}
    \State $D_K \gets \{X_i, M_{k,K}(y_i)\}_{i=1}^N$ \Comment{Updated dataset with increased dimension labels}
\EndProcedure

\Procedure{Dimension-Decreased Transformation}{$p(K), \rho_{K,y}$}
    \State $p \gets \text{WM}(p(K), \rho_{K,y})$ \Comment{Weighted mapping function to original dimension}
\EndProcedure

\Procedure{Gradient Normalization}{$\{g_b\}_{b=1}^B$}
    \State $\phi \gets \frac{1}{B} \sum_{b=1}^B \|g_b\|_2$ \Comment{Calculate mean l2-norm of gradients}
    \State $g_b' \gets \frac{g_b}{\|g_b\|_2} \cdot \phi$ \Comment{Normalize gradients}
\EndProcedure

\Procedure{Noise-Based Randomization}{$\tau, \mu$}
    \State $\gamma \sim \mathcal{N}(0, I_K)$ \Comment{Generate Gaussian noise}
    \State $\bar{\gamma} \gets \text{Softmax}(\gamma)$ \Comment{Normalize noise with Softmax}
    \State $\tau' \gets \tau \oplus (\mu \cdot \bar{\gamma})$ \Comment{Add noise to labels}
\EndProcedure

\Procedure{SecDT Training}{$D, K, \mu$}
    \State $D_K \gets \text{Dimension-Increased Transformation}(D, K)$
    \For{each mini-batch $\{g_b\}_{b=1}^B$}
        \State $\{g_b'\} \gets \text{Gradient Normalization}(\{g_b\}_{b=1}^B)$
    \EndFor
    \For{each sample's label $\tau$}
        \State $\tau' \gets \text{Noise-Based Randomization}(\tau, \mu)$
    \EndFor
    \State \textbf{Train} split learning models $E$ and $C$ using $D_K$
    \State \textbf{Output:} Models $E$ and $C$
\EndProcedure

\end{algorithmic}
\end{algorithm}
\section{Ablation Study}
\subsection{Impact of Expanded Dimension Size}\label{sec:eval_results:size}

\begin{figure*}[htbp]
  \centering
  \subfloat{\includegraphics[width=\textwidth]{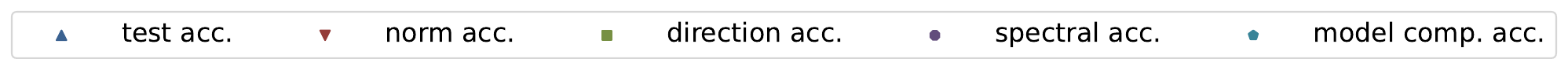}}
    \\
    \addtocounter{subfigure}{-1}
    \subfloat[{Avazu}]{\includegraphics[width=0.5\textwidth]{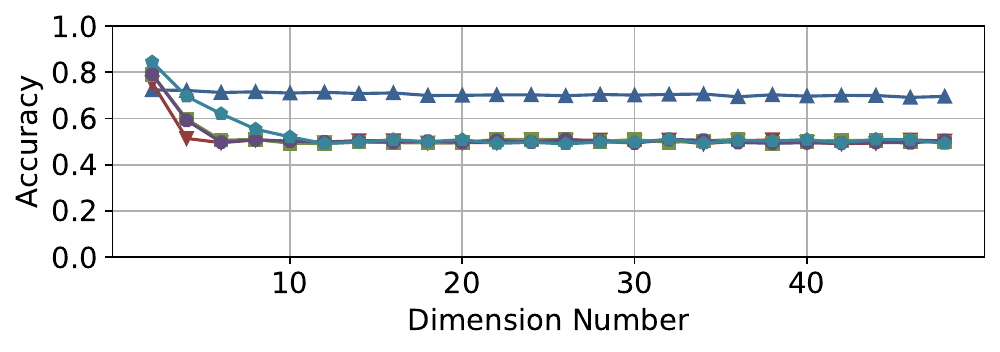}}
    \subfloat[{Criteo}]{\includegraphics[width=0.5\textwidth]{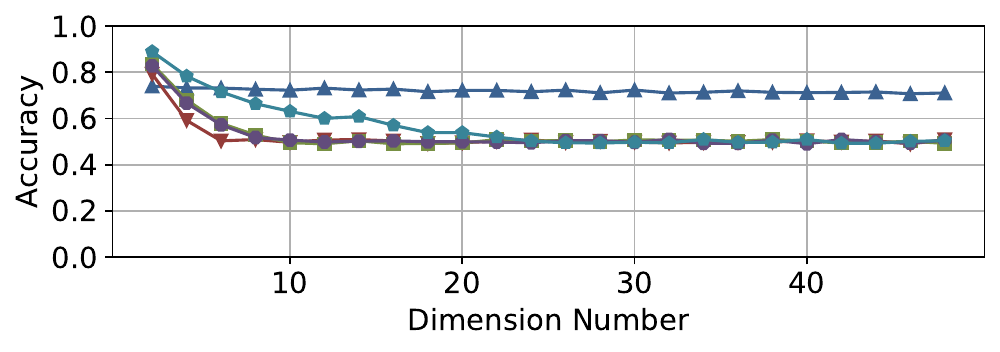}}
    \\
    \subfloat[{MNIST}]{\includegraphics[width=0.33\textwidth]{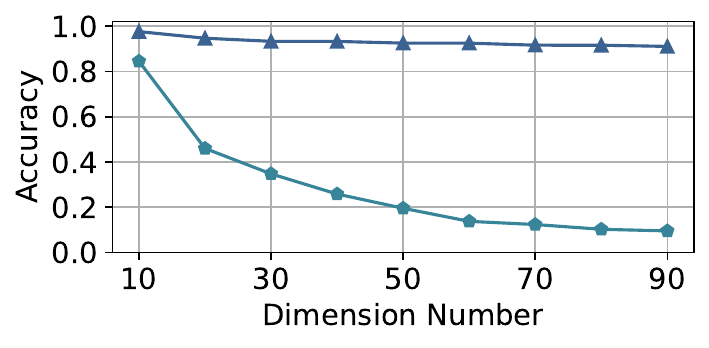}}
    \subfloat[{FashionMNIST}]{\includegraphics[width=0.33\textwidth]{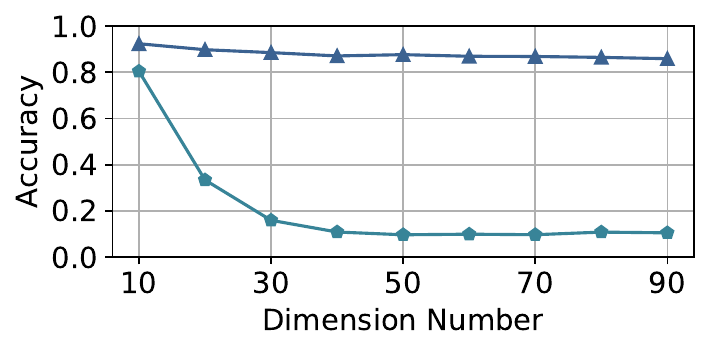}}
    \subfloat[{CIFAR}]{\includegraphics[width=0.33\textwidth]{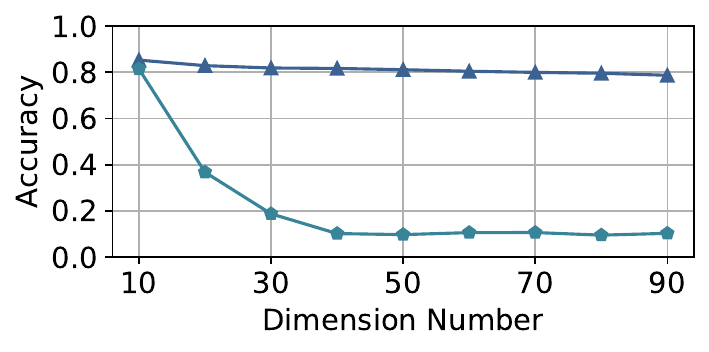}}
    \\
  \caption{
    Normalization plays a vital role in defending against norm and model completion attacks.
  }
  \label{fig:eval:dim_size}
\end{figure*}

\par We first evaluate the impact of expanded dimension size on the performance of split learning.
We applied all four attack schemes for the Avazu and Criteo datasets, while for MNIST, Fashion-MNIST, and CIFAR-10 datasets, we only applied the model competition attacks.

\par As the results shown in Figure \ref{fig:eval:dim_size}, the increasing dimension size does not affect the performance of the test utility.
While the attack utility is strongly affected as the dimension size increases.
For experiments on Avazu and Criteo, the norm, direction, and spectral attack utilization decrease soon and converge to the lower boundary as the dimension size reaches 8 (four times the original dimension size 2).
The model completion attack stood steady until the dimension size reached 20 (10 times the original dimension size).
This indicated that model completion attack requires \sysname to expand to a stronger dimension size to defend.
\subsection{Impact of Dimension Decrease}\label{sec:eval_results:decrease}

\begin{table*}
\centering
\begin{tabular}{M{2cm}|M{2.5cm}|M{2.5cm}|M{2.5cm}|M{2.5cm}}
\hline
{atk. type} &  norm attack &  direction attack  &   spectral attack &   mod. comp. \\ 
\hline\hline
\multicolumn{5}{c}{Criteo} \\ \hline
w/o decrease &  49.91/57.89 &  52.06/61.37 &  52.06/60.64 &   59.60/61.19 \\
decrease     &  52.34/76.95 &   55.57/78.10 &  55.17/77.48 &  62.87/76.55 \\\hline
\multicolumn{5}{c}{Avazu} \\ \hline
w/o decrease &  49.78/59.01 &   52.36/58.8 &   51.99/58.2 &  59.57/59.59 \\
decrease     &  52.52/76.94 &  54.92/74.77 &  55.23/76.32 &  63.42/75.42 \\\hline
\multicolumn{5}{c}{MNIST} \\ \hline
w/o decrease &       -/- &       -/- &      -/- &  16.06/72.43 \\
decrease     &       -/- &       -/- &      -/- &  16.94/97.61 \\\hline
\multicolumn{5}{c}{FashionMNIST} \\ \hline
w/o decrease &       -/- &       -/- &      -/- &  13.73/73.51 \\
decrease     &       -/- &       -/- &      -/- &  14.69/90.56 \\\hline
\multicolumn{5}{c}{CIFAR} \\ \hline
w/o decrease &       -/- &       -/- &      -/- &  12.54/66.5 \\
decrease     &       -/- &       -/- &      -/- &   13.3/84.80 \\\hline
\hline
\end{tabular}
\caption{
    We compared \sysname without dimension-decreased transformation and \sysname with dimension-decreased transformation.
    This table recorded the attack and test utility as "atk uti./test uti.".
    The results indicated that dimension-decreased transformation can maintain the accuracy performance of \sysname while not comprising \sysname's ability to defend various label inference attacks.
}
\label{tab:eval:dim_decrease}
\end{table*}

\par Furthermore, we compared \sysname without dimension-decreased transformation and \sysname with dimension-decreased transformation.
Dimension decrease is vital in keeping the accuracy of split learning when applying \sysname.

\par The dimension decrease transformation is a critical component of the \sysname framework that facilitates the conversion of the increased K-class labels back to the original binary or k-class labels. This process is essential for the practical application of the model, as it ensures that the output is usable for the end-user. As detailed in \S\ref{sec:scheme:dim}, \sysname employs a weighted mapping function to achieve this transformation, designed to preserve the model's performance while safeguarding against label inference attacks.

\par Our experiments were conducted on diverse datasets, including Avazu, Criteo, MNIST, FashionMNIST, and CIFAR-10, to evaluate the impact of the dimension decrease transformation. These datasets span various domains, comprehensively assessing the technique's effectiveness.

\par The experimental results, as depicted in Table \ref{tab:eval:dim_decrease} of the original document, offer profound insights into the significance of the dimension decrease transformation in \sysname. When \sysname was applied without incorporating the dimension decrease, a notable decline in test accuracy was observed across all evaluated datasets. This decline underscores the importance of the transformation in maintaining the model's predictive power.
Conversely, introducing the dimension decreased transformation and led to a substantial improvement in test accuracy. This enhancement was consistent across different datasets, demonstrating the technique's robustness. The results indicate that the weighted mapping function effectively navigates the increased dimensionality introduced during the training phase and accurately translates it back into the original label space.
Furthermore, the results highlight that the dimension decrease transformation does not compromise \sysname's ability to defend against label inference attacks. The attack utility remained consistently low, even when the transformation was applied, validating the technique's efficacy as a privacy-preserving measure.

\par The dimension decrease transformation's impact was also assessed against various attack vectors, including norm attacks, direction attacks, spectral attacks, and model completion attacks. The results demonstrated that the transformation did not adversely affect \sysname's resilience against these attacks. This finding is particularly significant, as it suggests that the dimension decrease can be universally applied to enhance the security of split learning models without diminishing their defense capabilities.

\par An additional observation from the experiments is the interaction between the noise level and the effectiveness of the dimension decrease transformation. While a moderate noise level did not significantly impact test utility, higher noise levels did lead to a decline in accuracy. This observation suggests that the dimension decrease transformation can counteract the adverse effects of noise to a certain extent, providing an additional layer of protection for the model.
\subsection{Impact of Normalization}\label{sec:eval_results:normalization}

\par We compared \sysname without normalization and \sysname with normalization.
Normalization is vital in defending against norm attacks when applying \sysname.

\par \sysname's approach to gradient normalization, as detailed in \S\ref{sec:scheme:norm}, involves using the mean $l2$-norm of the gradients within a mini-batch as a standard for normalization. This method is particularly effective against attacks that rely on the magnitude of gradients to infer sensitive information. By enforcing a standard norm, \sysname diminishes the adversaries' ability to discern differences in gradient magnitudes that could indicate the presence of certain classes within the dataset.

\begin{table*}
\centering
\begin{tabular}{M{2cm}|M{2.5cm}|M{2.5cm}|M{2.5cm}|M{2.5cm}}
\hline
{atk. type} &  norm attack &  direction attack  &   spectral attack &   mod. comp. \\ 
\hline\hline
\multicolumn{5}{c}{Criteo} \\ \hline
w/o random. &  57.31/77.78 &  65.66/76.59 &  65.50/77.64 &  70.47/76.84 \\
random.     &  57.32/74.87 &  53.37/76.07 &  53.77/75.90 &  63.28/75.80 \\\hline
\multicolumn{5}{c}{Avazu} \\ \hline
w/o random. &  57.84/74.83 &  66.42/76.36 &  65.58/76.17 &  71.61/75.22 \\
random.     &  57.77/73.72 &  53.22/73.85 &  53.27/75.09 &  62.76/74.85 \\\hline
\multicolumn{5}{c}{MNIST} \\ \hline
w/o random. &       -/- &       -/- &      -/- &  28.26/96.22 \\
random.     &       -/- &       -/- &      -/- &  17.05/94.31 \\\hline
\multicolumn{5}{c}{FashionMNIST} \\ \hline
w/o random. &       -/- &       -/- &      -/- &  24.99/93.24 \\
random.     &       -/- &       -/- &      -/- &  14.41/90.21 \\\hline
\multicolumn{5}{c}{CIFAR} \\ \hline
w/o random. &       -/- &       -/- &      -/- &  22.37/85.62 \\
random.     &       -/- &       -/- &      -/- &  13.26/81.73 \\\hline
\hline
\end{tabular}
\caption{
    We compared \sysname without normalization and \sysname with normalization.
    This table recorded the attack and test utility as "atk uti./test uti.".
    The results indicated that normalization can maintain the accuracy of \sysname while not comprising \sysname's ability to defend various label inference attacks.
    Normalization is vital in defending against norm and model completion attacks.
}
\label{tab:eval:normalization}
\end{table*}

\par The use of gradient normalization in \sysname introduces a trade-off between utility and privacy. While normalization can reduce the model's utility for adversaries, it may also affect the model's ability to learn complex patterns if not implemented carefully. However, our experiments, as shown in Table \ref{tab:eval:normalization}, demonstrate that \sysname successfully strikes a balance, maintaining high test accuracy while significantly reducing the attack utility.
\subsection{Impact of Noise}\label{sec:eval_results:noise}

\par Finally, we evaluate the impact of noise scale on the performance of split learning.
We applied all four attack schemes for the Avazu and Criteo datasets, while we only applied the model competition attacks for the MNIST, Fashion-MNIST, and CIFAR-10 datasets.

\par The incorporation of noise into the learning process is a fundamental strategy for enhancing privacy in machine learning, particularly in the context of split learning, where protecting sensitive labels is paramount. As outlined in \S\ref{sec:scheme:noise}, \sysname introduces noise-based randomization to obfuscate the mapping pools and to prevent adversaries from discerning the true dimensionality of the label space (K). This approach complements the dimension transformation and gradient normalization techniques within \sysname, offering a layered defense against label inference attacks.

\par Our experiments were structured to meticulously evaluate the impact of noise-based randomization on the performance of \sysname. A spectrum of noise levels was applied to the label encoding process, and the outcomes were assessed against a backdrop of real-world datasets, including Avazu, Criteo, MNIST, FashionMNIST, and CIFAR-10. The objective was to determine how varying noise levels affect the model's utility for legitimate tasks and its resilience against adversarial attacks.

\begin{figure*}[htbp]
  \centering
  \subfloat{\includegraphics[width=\textwidth]{Figures/evaluation_figures/2-dim_size/2-dim_size_legend.pdf}}
    \\
    \addtocounter{subfigure}{-1}
    \subfloat[{Avazu}]{\includegraphics[width=0.5\textwidth]{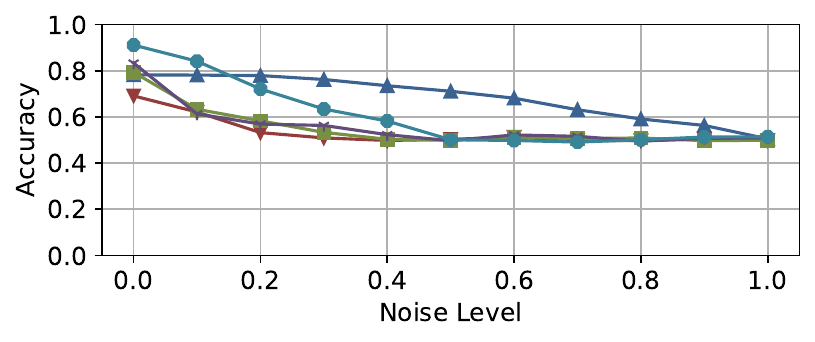}}
    \subfloat[{Criteo}]{\includegraphics[width=0.5\textwidth]{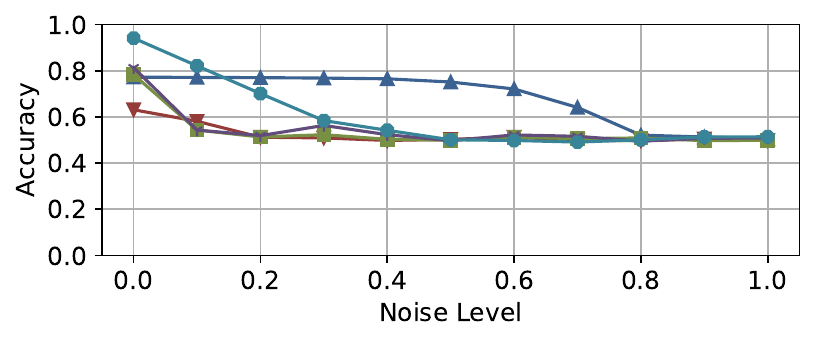}}
    \\
    \subfloat[{MNIST}]{\includegraphics[width=0.33\textwidth]{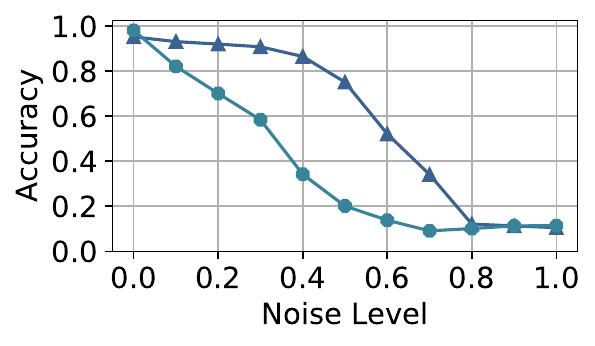}}
    \subfloat[{FashionMNIST}]{\includegraphics[width=0.33\textwidth]{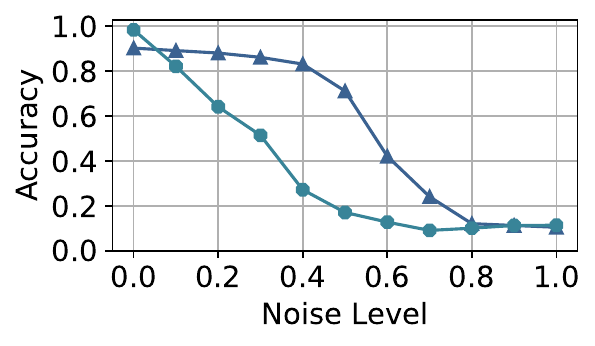}}
    \subfloat[{CIFAR}]{\includegraphics[width=0.33\textwidth]{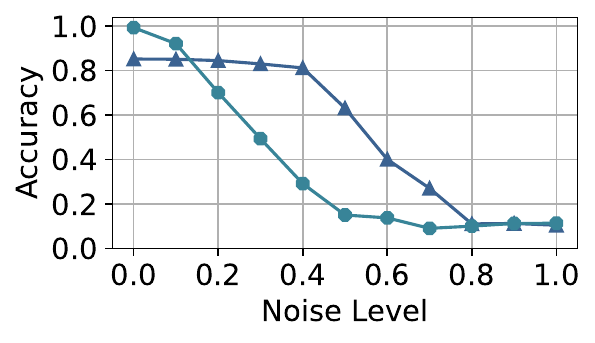}}
    \\
  \caption{
    The evaluation results of noise's impact on \sysname's performance.
  }
  \label{fig:eval:noise}
\end{figure*}

\par The experimental findings, as presented in Figure \ref{fig:eval:noise} of the original document, offer a nuanced perspective on the role of noise in \sysname. A key observation was that the introduction of noise did not significantly impair the model's performance on test tasks up to a certain threshold. This suggests that the noise was effectively integrated without disrupting the model's learning capabilities.
However, test utility markedly declined beyond a noise level of 0.5. This decline indicates a critical threshold beyond which the noise erodes the model's ability to make accurate predictions. Identifying this threshold is crucial for practitioners seeking to balance privacy and utility in their models.

\par In contrast to the impact on test utility, the introduction of noise had a pronounced detrimental effect on attack utility. The results demonstrated that the model's utility for adversarial purposes decreased significantly as the noise level increased. Notably, for norm, direction, and spectral attacks, the attack utility plummeted as the noise scale surpassed 0.2, effectively neutralizing these attack vectors at moderate noise levels.

\par An interesting observation was the resilience of model completion attacks to noise up to a level of 0.4. This suggests that adversaries employing this type of attack may require a higher noise threshold to be deterred. The persistence of model completion attacks at lower noise levels underscores the need for a multi-faceted defense strategy, where noise is complemented by other techniques such as dimension transformation and gradient normalization.
\section{Inference Attack towards $K$ in \sysname}\label{sec:eval_results:attack}

\begin{figure*}
  \centering
    \subfloat[$K/k=2$]{\includegraphics[width=0.33\textwidth]{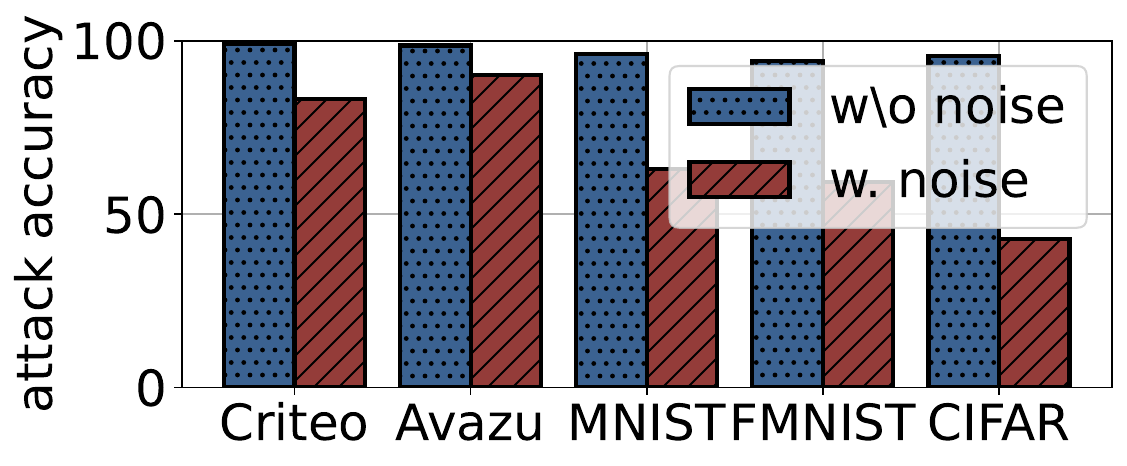}}
    \subfloat[$K/k=5$]{\includegraphics[width=0.33\textwidth]{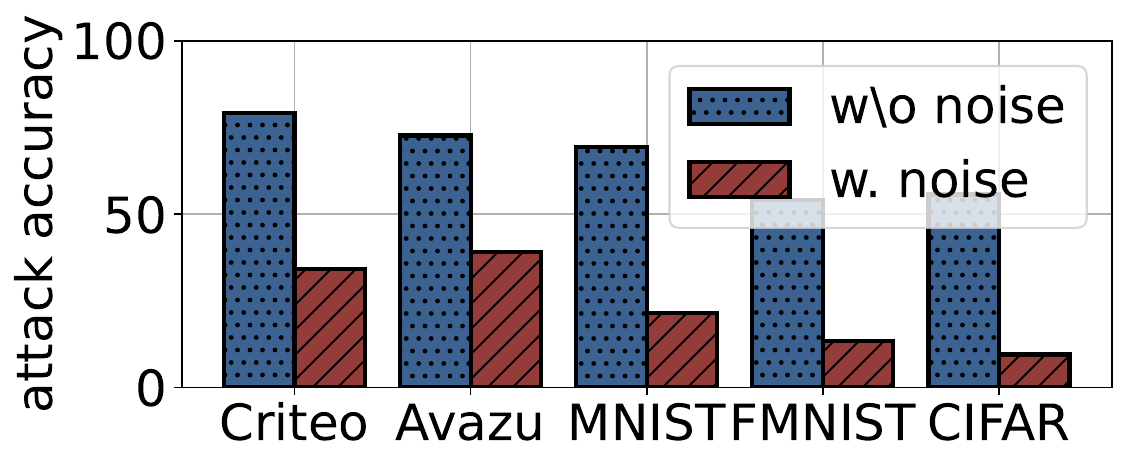}}
    \subfloat[$K/k=10$]{\includegraphics[width=0.33\textwidth]{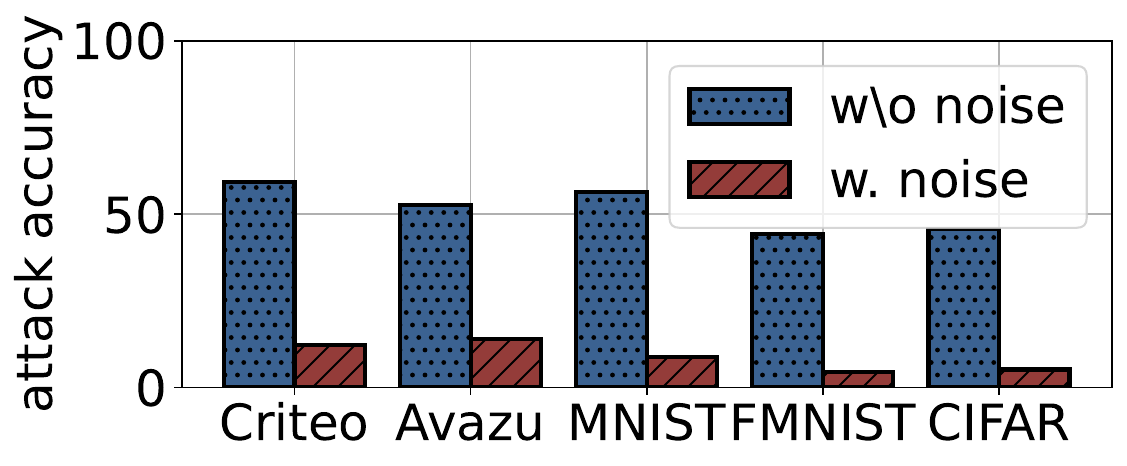}}
    \\
  \caption{
  The attack accuracy of a potential attacker using K-means based method for inferring the value of $K$
  }
  \label{fig:eval:kmeans}
\end{figure*}

\par When an attacker knows our \sysname is used to learn the global model, it can adapt its attacks to \sysname. The attacker can guess the increased dimension $K$ in \sysname. Since the mapping from 2-dimension to $K$-dimension $\mathcal{M}_{2, K}$ will reveal some clustering features. We evaluate whether the attacker can guess the real value of $K$ in \sysname. For each data sample's backward gradient $g$, we iteratively set the dimension size from 2 to $2K$. 
Then, we use the K-means clustering algorithm to cluster the gradient to the set dimension number of clusters. After that, the calinski harabasz score is used to evaluate the clustering result. The set dimension with the highest will be elected as the guessed dimension. 
We set the value of $K/k$ to be 2, 5, and 10 in our experiment.

\par We evaluate \sysname w/ and w/o noise in Figure \ref{fig:eval:kmeans}, and the result shows that the attacker can easily guess the dimension $K$ of \sysname since the frequency it guesses right is very high. Furthermore, we evaluate \sysname with our proposed SGN and find out that the frequency that the attacker guesses the right dimension $K$ becomes very low. Thus, the result indicates that our proposed noise can ease the proposed adaptive attack toward \sysname. 
\section{Time Cost}\label{sec:eval_results:cost}

\begin{figure*}[htbp]
  \centering
    \subfloat[{Avazu}]{\includegraphics[width=0.5\textwidth]{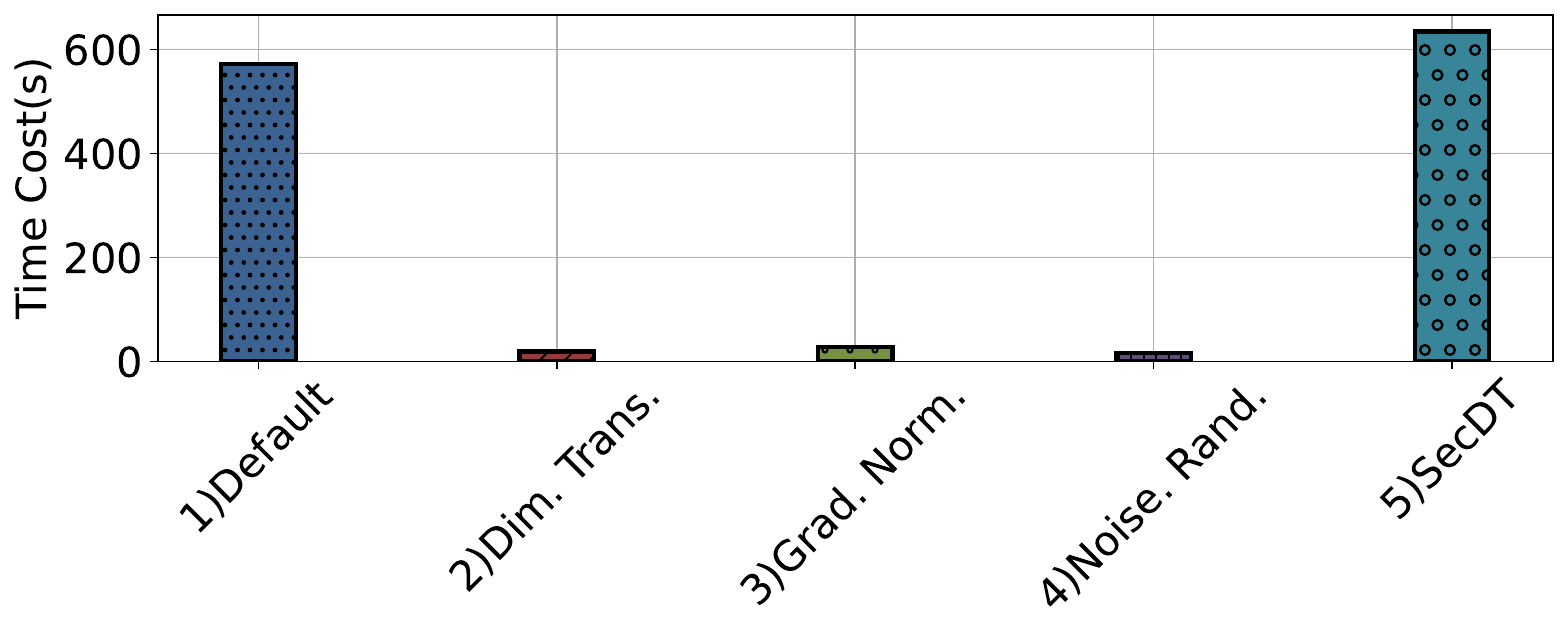}}
    \subfloat[{Criteo}]{\includegraphics[width=0.5\textwidth]{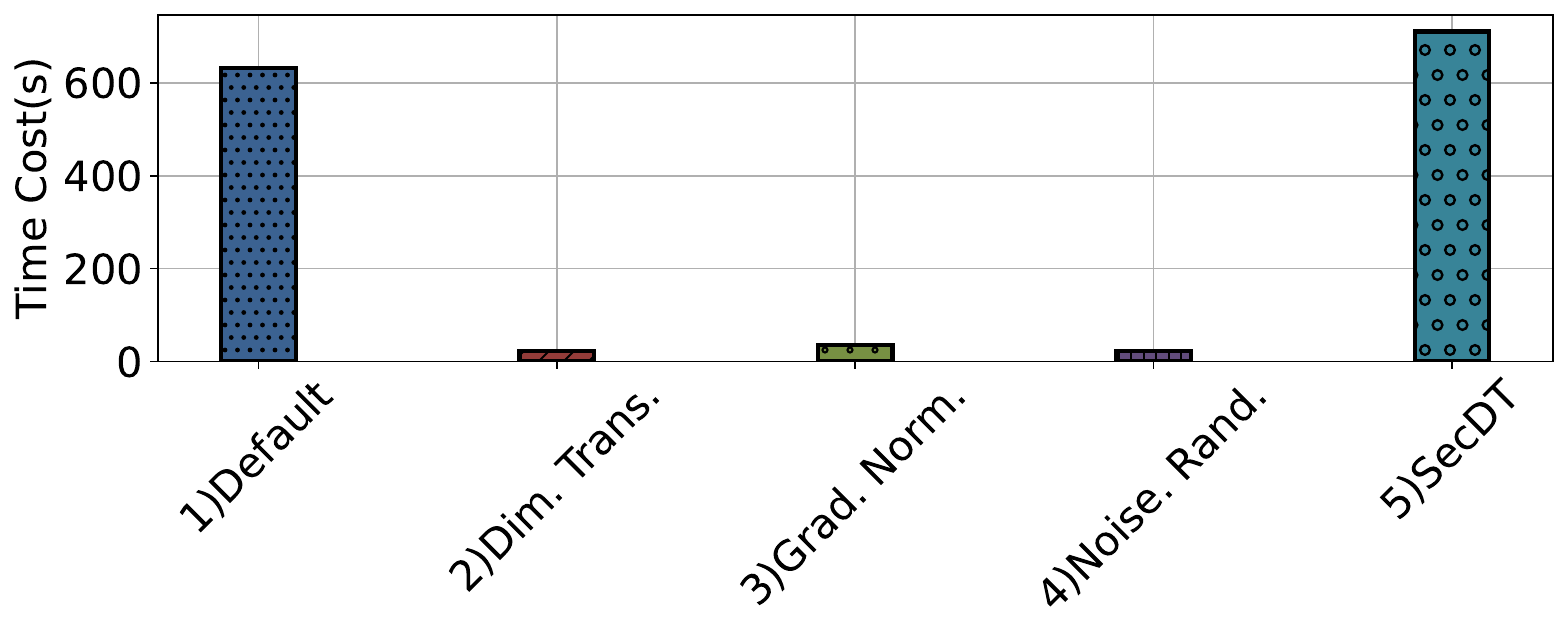}}
    \\
    \vspace{-10pt}
    \subfloat[MNIST]{\includegraphics[width=0.33\textwidth]{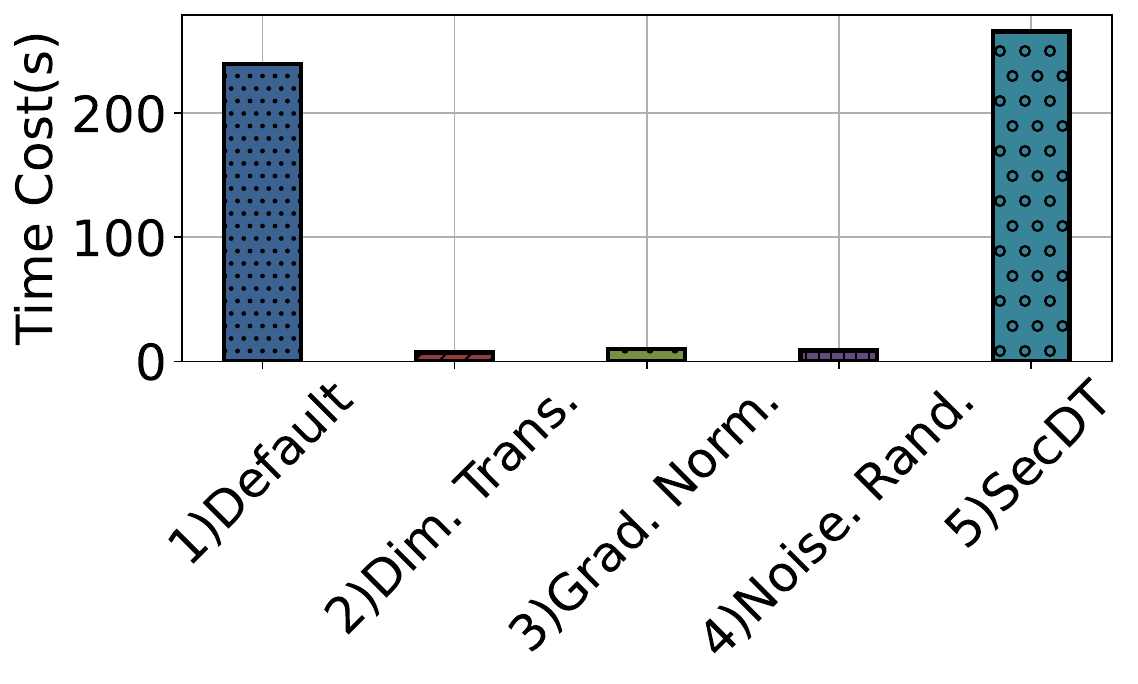}}
    \subfloat[{FashionMNIST}]{\includegraphics[width=0.33\textwidth]{Figures/evaluation_figures/8-cost/8-cost-MNIST.pdf}}
    \subfloat[{CIFAR}]{\includegraphics[width=0.33\textwidth]{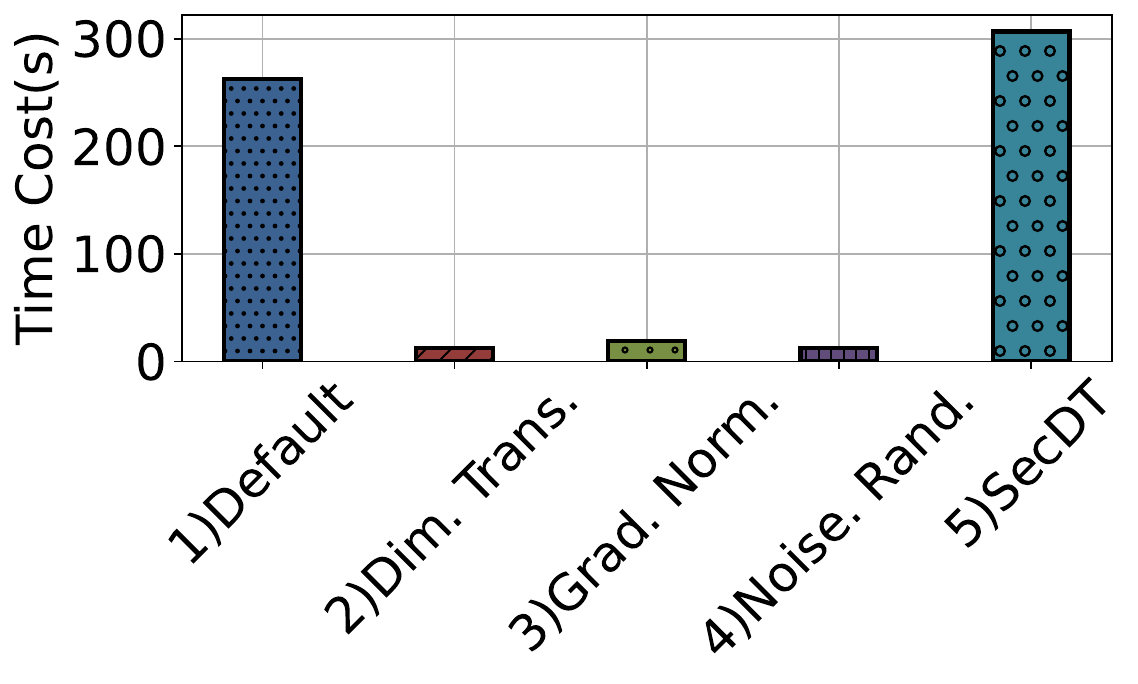}}
    \\
  \caption{
    This figure evaluates \sysname's time cost on five evaluated datasets for different processes: 
    \textbf{1) Default}: The default split learning without any defense.
    \textbf{2) Dim. Trans.}: The dimension transformation process of \sysname.
    \textbf{3) Grad. Norm.}: The gradient normalization process of \sysname.
    \textbf{4) Noise. Rand.}: The noise-based randomization process of \sysname.
    \textbf{5) \sysname}: The whole process of \sysname (including 1)+2)+3)+4)).
    The evaluation results show that the time cost of \sysname is not greatly increased compared with default split learning.
  }
  \label{fig:eval:cost}
\end{figure*}

To conduct a comprehensive assessment, we designed experiments that measured the time expenditure associated with various processes within \sysname, including dimension transformation, gradient normalization, and noise-based randomization steps. These individual processes were evaluated separately and in aggregate to provide a clear picture of their contributions to the overall time cost.
Our experimental setup involved a series of controlled tests on diverse datasets, such as Avazu, Criteo, MNIST, FashionMNIST, and CIFAR, to ensure the generalizability of our findings. Figure \ref{fig:eval:cost} shows that integrating \sysname into the split learning framework results in a modest increase in time cost, approximately 11\%, relative to the baseline split learning model without any defense mechanisms.

This moderate increase in time cost can be attributed to the additional computational steps required to perform secure dimension transformation, gradient normalization, and the introduction of noise randomization. Specifically, the dimension transformation process involves mapping the original labels to a higher-dimensional space, which, while enhancing privacy, necessitates extra computational effort. Similarly, gradient normalization requires the computation of l2-norms for each gradient in the mini-batch, followed by a scaling operation to standardize these gradients. Lastly, the noise-based randomization step injects an element of randomness into the label space, further complicating the learning process and contributing to the overall time cost.

It is important to note that the observed time overhead is a reasonable trade-off for the enhanced security and privacy protections offered by \sysname. The relatively small increase in time cost suggests that \sysname is a scalable solution that can be feasibly implemented in various split-learning applications.
\section{Proof of Convergence}

First, we made the following assumptions:
\begin{enumerate}
    \item The loss function $L(\theta)$ is Lipschitz continuous, that is, there exists a constant $L > 0$ such that for any $\theta_1$ and $\theta_2$, we have $|L(\theta_1) - L(\theta_2)| \leq L \|\theta_1 - \theta_2\| $
    \item The gradient $\nabla L(\theta)$ is bounded, that is, there exists a constant $G > 0$ such that for any $\theta$, we have $\|\nabla L(\theta)\| \leq G$
    \item The learning rate $\eta_t$ satisfies the standard learning rate condition, that is, $\sum_{t=1}^{\infty} \eta_t = \infty \quad \text{and} \quad \sum_{t=1}^{\infty} \eta_t^2 < \infty$
\end{enumerate}

Next, we will analyze the impact of each step of SecDT on the loss function step by step and finally prove its convergence.

\sssec{Dimension conversion.} During the dimension conversion process, we convert the original k-dimensional label $y_i$ into a K-dimensional label $M_{k, K}(y_i)$. Assume that the original loss function is $L(\theta; X, y)$ and the converted loss function is $L'(\theta; X, M_{k, K}(y))$. We first analyze the change in the loss function.

Since the dimension conversion is only an expansion of the label space and does not affect the expressiveness of the feature space and the model, there is a fixed relationship between the original loss function and the converted loss function for a fixed model parameter $\theta$. Specifically, assuming that $\hat{y}$ is the predicted value of the original label and $\hat{y}'$ is the predicted value of the label after dimension conversion, we have:
\begin{equation}
L(\theta; X, y) = L'(\theta; X, M_{k,K}(y))
\end{equation}

\sssec{Noise Randomization.} In the noise randomization step, we add noise $\gamma$ to the label to obtain a new label $\tau' = \tau \oplus (\mu \cdot \bar{\gamma})$. The loss function after adding noise is $L''(\theta; X, \tau')$. We need to analyze the impact of noise on the loss function.

Since the noise $\gamma$ follows a standard Gaussian distribution and is normalized by the Softmax function, its amplitude is controllable. Assume the intensity of the noise is $\mu$, then for any $\theta$, we have:
\begin{equation}
L'(\theta; X, M_{k,K}(y)) \leq L''(\theta; X, \tau') + \mu \cdot \|\gamma\|
\end{equation}
According to Gaussian noise's properties, the noise's effect $\gamma$ can be regarded as a constant term.

\sssec{Gradient Normalization.} In the gradient normalization step, we normalize each gradient. Assume the normalized gradient is $\bar{g}_b$, then:
\begin{equation}
\bar{g}_b = g_b \cdot \frac{\phi}{\|g_b\|}
\end{equation}
Where $\phi$ is the mean $l_2$-norm of the gradient.

The purpose of gradient normalization is to balance the gradient amplitudes of different categories, thereby reducing the impact of the gradient amplitude on the loss function. Since the normalization operation of the gradient is linear, there is a fixed proportional relationship between the normalized gradient and the original gradient. That is, for any $\theta$ and gradient $\nabla L(\theta)$, there is: \begin{equation}\nabla L(\theta) \approx \bar{g}_b\end{equation}

\sssec{Convergence Proof.} We consider the entire training process of SecDT. Assume $\theta_t$ is the model parameter of the $t$th iteration, $\eta_t$ is the learning rate, and $\nabla L(\theta_t)$ is the current gradient, then the update rule is:
\begin{equation}
\theta_{t+1} = \theta_t - \eta_t \bar{g}_b
\end{equation}

By the Lipschitz continuity assumption, for any $\theta_t$ and $\theta_{t+1}$, we have:
\begin{equation}
L(\theta_{t+1}) \leq L(\theta_t) - \eta_t \nabla L(\theta_t) \cdot \bar{g}_b + \frac{L}{2} \eta_t^2 \|\bar{g}_b\|^2
\end{equation}

Since $\bar{g}_b$ is the normalized gradient, whose norm is bounded, so there exists a constant $C > 0$ such that $\|\bar{g}_b\| \leq C$. Therefore, the above formula can be rewritten as:
\begin{equation}
L(\theta_{t+1}) \leq L(\theta_t) - \eta_t \|\nabla L(\theta_t)\|^2 + \frac{L}{2} \eta_t^2 C^2
\end{equation}

Summing the above formula and using the learning rate condition, we get:
\begin{equation}
\sum_{t=1}^{T} \eta_t \|\nabla L(\theta_t)\|^2 \leq L(\theta_1) - L(\theta_{T+1}) + \frac{L}{2} C^2 \sum_{t=1}^{T} \eta_t^2
\end{equation}

Since $\sum_{t=1}^{\infty} \eta_t^2 < \infty$, the sum on the right is bounded, let its upper bound be $B$, then:
\begin{equation}
\sum_{t=1}^{T} \eta_t \|\nabla L(\theta_t)\|^2 \leq L(\theta_1) - L(\theta_{T+1}) + B
\end{equation}

Since $\sum_{t=1}^{\infty} \eta_t = \infty$, in order to ensure the left-hand side converges, $\|\nabla L(\theta_t)\|$ must approach zero. This means that the loss function $L(\theta_t)$ will converge to a bounded range.

In summary, we have proved that during the training of SecDT, the loss function $L(\theta)$ will converge to a bounded range, that is, there is a constant $M > 0$ such that:
\begin{equation}
\lim_{t \to \infty} L(\theta_t) \leq M
\end{equation}

In order to determine the specific upper bound $M$, we need to analyze the convergence behavior of the above inequality term by term in detail. When $T \to \infty$, the sum of $\sum_{t=1}^{T} \eta_t \|\nabla L(\theta_t)\|^2$ will tend to $L(\theta_1) + B$.

Assume that the sum of $\|\nabla L(\theta_t)\|^2$ approaches a certain limit value $S$, then:

\begin{equation}
S = \sum_{t=1}^{\infty} \eta_t \|\nabla L(\theta_t)\|^2
\end{equation}

We can express the final upper bound $M$ as:

\begin{equation}
M = L(\theta_1) + B - S
\end{equation}

Since $S$ represents the infinite sum of the squared sum of gradients, and according to $\sum_{t=1}^{\infty} \eta_t = \infty$ and $\sum_{t=1}^{\infty} \eta_t^2 < \infty$, we can further determine the specific value through the learning rate and gradient convergence behavior in the specific gradient descent algorithm.

\sssec{Conclusion} Based on the above analysis, we can determine the specific value of the constant $M$ as:

\begin{equation}
M = L(\theta_1) + \frac{L}{2} C^2 \sum_{t=1}^{\infty} \eta_t^2 - \sum_{t=1}^{\infty} \eta_t \|\nabla L(\theta_t)\|^2
\end{equation}

Where:
\begin{itemize}
    \item $L(\theta_1)$ is the initial loss value
    \item $\frac{L}{2} C^2 \sum_{t=1}^{\infty} \eta_t^2$ is the upper bound of the noise effect
    \item $\sum_{t=1}^{\infty} \eta_t \|\nabla L(\theta_t)\|^2$ is the sum of the loss reduction during the gradient descent process
\end{itemize}

\end{document}